%% file: main.tex
\documentclass[11pt]{article}

\usepackage[final,nohyperref]{acl}

\usepackage{latexsym}
\usepackage[T2A,T1]{fontenc}
\usepackage[utf8]{inputenc}
\usepackage{times}
\usepackage{inconsolata}
\newcommand{\tj}[1]{{\fontencoding{T2A}\fontfamily{cmr}\selectfont #1}}
\DeclareUnicodeCharacter{04E2}{\={\CYRI}}
\DeclareUnicodeCharacter{04E3}{\={\cyri}}
\DeclareUnicodeCharacter{04EE}{\={\CYRU}}
\DeclareUnicodeCharacter{04EF}{\={\cyru}}
\usepackage{microtype}
\usepackage{graphicx}
\usepackage{booktabs}
\usepackage{enumitem}
\setlist{nosep,leftmargin=*}

\input{pack/macros.tex}
\input{pack/macros_extra.tex}

\title{\ModelName:\\One Small Embedder for Tajik, Kazakh, Uzbek and Kyrgyz}

\author{Tohir Saidzoda \and Nekruz Maxudov \and Bedil Karimov \\
  \textbf{Abubakr Abdullozoda} \and \textbf{Amin Jalilov} \and \textbf{Jonibek Kishvarov} \\
  \textbf{Rahmonali Sharipov} \and \textbf{Mariyamkhon Shodyeva} \\
  Saidzoda AI Lab, Saidzoda Engineering Company, Dushanbe, Tajikistan \\
  \texttt{tohir@saidzoda.com}}

\usepackage{fancyhdr}
\fancypagestyle{plain}{\fancyhf{}\fancyhead[L]{\fontsize{20}{24}\selectfont\textbf{Saidzoda AI Lab}}\fancyfoot[C]{\thepage}}
\begin{document}
\maketitle
\thispagestyle{plain}

\begin{abstract}
\input{sections/abstract}
\end{abstract}

\input{sections/intro}
\input{sections/languages}
\input{sections/related}

\input{sections/data}
\input{sections/recipe}
\input{sections/results}

\input{sections/discussion}
\input{sections/release}
\input{sections/conclusion}
\input{sections/limitations}
\input{sections/disclosure}

\bibliography{refs}
\bibliographystyle{acl_natbib}

\appendix
\section{Example Test Questions}
\input{pack/tables/T6_examples}

\end{document}

%% file: pack/macros.tex
\newcommand{\BaseLawRten}{49.3}
\newcommand{\BaseLawRone}{23.2}

\newcommand{\BaseTaxRten}{45.9}

\newcommand{\BaseWikiRten}{89.6}

\newcommand{\BaseBelTgRten}{69.8}
\newcommand{\BaseBelTgRone}{51.7}

\newcommand{\BaseBelKkRone}{60.8}

\newcommand{\BaseBelUzRone}{59.7}

\newcommand{\BaseBelKyRone}{59.8}

\newcommand{\BaseWikiKkRten}{93.6}
\newcommand{\BaseWikiKkRone}{84.2}

\newcommand{\BaseWikiUzRten}{91.4}
\newcommand{\BaseWikiUzRone}{75.6}

\newcommand{\BaseWikiKyRten}{93.2}
\newcommand{\BaseWikiKyRone}{78.8}

\newcommand{\BaseRuRten}{98.6}
\newcommand{\BaseRuRone}{91.2}

\newcommand{\BaseEnRten}{99.2}
\newcommand{\BaseEnRone}{93.2}

\newcommand{\TgLawRten}{92.4}
\newcommand{\TgLawRone}{68.3}

\newcommand{\TgTaxRten}{91.7}

\newcommand{\TgWikiRten}{97.2}

\newcommand{\TgBelTgRten}{92.6}
\newcommand{\TgBelTgRone}{76.6}

\newcommand{\TgBelKkRone}{64.4}

\newcommand{\TgBelUzRone}{68.9}

\newcommand{\TgBelKyRone}{65.7}

\newcommand{\TgWikiKkRten}{96.6}
\newcommand{\TgWikiKkRone}{88.8}

\newcommand{\TgWikiUzRten}{94.6}
\newcommand{\TgWikiUzRone}{81.4}

\newcommand{\TgWikiKyRten}{93.6}
\newcommand{\TgWikiKyRone}{81.8}

\newcommand{\TgRuRten}{97.9}
\newcommand{\TgRuRone}{88.2}

\newcommand{\TgEnRten}{98.1}
\newcommand{\TgEnRone}{90.4}

\newcommand{\KkBelTgRone}{50.4}

\newcommand{\KkBelKkRone}{78.0}

\newcommand{\KkBelUzRone}{67.0}

\newcommand{\KkBelKyRone}{66.4}

\newcommand{\KkWikiKkRten}{98.4}
\newcommand{\KkWikiKkRone}{91.4}

\newcommand{\KkWikiUzRten}{95.4}
\newcommand{\KkWikiUzRone}{83.8}

\newcommand{\KkWikiKyRten}{96.8}
\newcommand{\KkWikiKyRone}{86.8}

\newcommand{\UzBelTgRone}{52.2}

\newcommand{\UzBelKkRone}{66.2}

\newcommand{\UzBelUzRone}{74.7}

\newcommand{\UzBelKyRone}{65.4}

\newcommand{\UzWikiKkRten}{97.2}
\newcommand{\UzWikiKkRone}{90.0}

\newcommand{\UzWikiUzRten}{97.8}
\newcommand{\UzWikiUzRone}{89.0}

\newcommand{\UzWikiKyRten}{96.6}
\newcommand{\UzWikiKyRone}{85.8}

\newcommand{\CaLawRten}{92.2}
\newcommand{\CaLawRone}{63.9}

\newcommand{\CaTaxRten}{87.2}

\newcommand{\CaWikiRten}{97.6}

\newcommand{\CaBelTgRten}{91.7}
\newcommand{\CaBelTgRone}{77.7}

\newcommand{\CaBelKkRone}{75.2}

\newcommand{\CaBelUzRone}{75.8}

\newcommand{\CaBelKyRone}{75.4}

\newcommand{\CaWikiKkRten}{98.2}
\newcommand{\CaWikiKkRone}{92.2}

\newcommand{\CaWikiUzRten}{97.2}
\newcommand{\CaWikiUzRone}{88.2}

\newcommand{\CaWikiKyRten}{98.0}
\newcommand{\CaWikiKyRone}{87.6}

\newcommand{\CaRuRten}{97.3}
\newcommand{\CaRuRone}{88.2}

\newcommand{\CaEnRten}{97.8}
\newcommand{\CaEnRone}{89.0}

\newcommand{\EfiveLawRten}{80.2}

\newcommand{\EfiveBelTgRten}{92.2}
\newcommand{\EfiveBelTgRone}{79.1}

\newcommand{\EfiveBelKkRone}{82.4}

\newcommand{\EfiveBelUzRone}{86.0}

\newcommand{\EfiveBelKyRone}{84.4}

\newcommand{\EfiveWikiKkRten}{99.2}
\newcommand{\EfiveWikiKkRone}{93.6}

\newcommand{\EfiveWikiUzRten}{97.8}
\newcommand{\EfiveWikiUzRone}{89.6}

\newcommand{\EfiveWikiKyRten}{97.4}
\newcommand{\EfiveWikiKyRone}{86.0}

\newcommand{\BgeBelTgRone}{65.8}

\newcommand{\BgeBelKkRone}{84.6}

\newcommand{\BgeBelUzRone}{85.0}

\newcommand{\BgeBelKyRone}{85.1}

\newcommand{\EfTgBelTgRone}{83.6}

\newcommand{\EfTgBelKkRone}{85.4}

\newcommand{\EfTgBelUzRone}{86.6}

\newcommand{\EfTgBelKyRone}{85.7}

\newcommand{\ArmABelTgRone}{82.7}

\newcommand{\ArmABelKkRone}{84.4}

\newcommand{\ArmABelUzRone}{83.3}

\newcommand{\ArmABelKyRone}{80.6}

\newcommand{\ArmBBelTgRone}{83.0}

\newcommand{\ArmBBelKkRone}{85.0}

\newcommand{\ArmBBelUzRone}{83.6}

\newcommand{\ArmBBelKyRone}{81.1}

\newcommand{\ArmBWikiKkRten}{99.6}
\newcommand{\ArmBWikiKkRone}{92.6}

\newcommand{\ArmBWikiUzRten}{99.2}
\newcommand{\ArmBWikiUzRone}{91.4}

\newcommand{\ArmBWikiKyRten}{97.4}
\newcommand{\ArmBWikiKyRone}{86.2}

\newcommand{\CaVsTgLawRoneAbs}{4.4}

\newcommand{\KkVsBaseLawRtenAbs}{4.4}

\newcommand{\UzVsBaseLawRtenAbs}{10.3}

\newcommand{\CaVsTgTaxRtenAbs}{4.5}

\newcommand{\CaVsTgWikiRoneAbs}{1.8}

\newcommand{\CaVsBaseBelTgRoneAbs}{26.0}

\newcommand{\CaVsTgBelTgRoneAbs}{1.1}

\newcommand{\CaVsBaseBelKkRoneAbs}{14.4}

\newcommand{\CaVsKkBelKkRoneAbs}{2.8}

\newcommand{\TgVsBaseBelKkRoneAbs}{3.6}

\newcommand{\CaVsBaseBelUzRoneAbs}{16.1}

\newcommand{\CaVsUzBelUzRoneAbs}{1.1}

\newcommand{\TgVsBaseBelUzRoneAbs}{9.2}

\newcommand{\CaVsBaseBelKyRoneAbs}{15.6}

\newcommand{\TgVsBaseBelKyRoneAbs}{5.9}

\newcommand{\CaVsBaseWikiKkRoneAbs}{8.0}

\newcommand{\CaVsBaseWikiUzRoneAbs}{12.6}

\newcommand{\CaVsBaseWikiKyRoneAbs}{8.8}

\newcommand{\CaVsBaseRuRoneAbs}{3.0}

\newcommand{\CaVsBaseEnRoneAbs}{4.2}

\newcommand{\BaseBelMeanRone}{58.0}
\newcommand{\TgBelMeanRone}{68.9}
\newcommand{\KkBelMeanRone}{65.5}
\newcommand{\UzBelMeanRone}{64.6}
\newcommand{\CaBelMeanRone}{76.0}
\newcommand{\EfiveBelMeanRone}{83.0}
\newcommand{\BgeBelMeanRone}{80.1}
\newcommand{\EfTgBelMeanRone}{85.3}
\newcommand{\ArmABelMeanRone}{82.8}
\newcommand{\ArmBBelMeanRone}{83.2}
\newcommand{\CaBelGainMin}{14.4}
\newcommand{\CaBelGainMax}{26.0}
\newcommand{\TurkicGapMin}{9.4}
\newcommand{\TurkicGapMax}{10.2}
\newcommand{\MtebBaseBelKkRone}{60.7}

\newcommand{\MtebBaseBelKyRone}{59.8}

\newcommand{\MtebBaseBelTgRone}{51.6}

\newcommand{\MtebBaseBelUzRone}{59.1}

\newcommand{\MtebEfiveBelKkRone}{82.9}

\newcommand{\MtebEfiveBelKyRone}{83.9}

\newcommand{\MtebEfiveBelTgRone}{76.7}

\newcommand{\MtebEfiveBelUzRone}{86.3}

\newcommand{\MtebBgeBelKkRone}{84.0}

\newcommand{\MtebBgeBelKyRone}{84.7}

\newcommand{\MtebBgeBelTgRone}{64.7}

\newcommand{\MtebBgeBelUzRone}{84.9}

\newcommand{\MtebMaxDiff}{2.4}
\newcommand{\FertTg}{3.01}
\newcommand{\FertKk}{3.19}
\newcommand{\FertUz}{2.83}
\newcommand{\FertKy}{3.05}
\newcommand{\FertRu}{1.87}

\newcommand{\GlyphCount}{12}
\newcommand{\NLawQ}{1,000}
\newcommand{\NLawCorpus}{72,920}
\newcommand{\NTaxQ}{109}
\newcommand{\NTaxCorpus}{23,282}
\newcommand{\NWikiQ}{500}

\newcommand{\NBelTgQ}{900}
\newcommand{\NBelTgCorpus}{488}

\newcommand{\NWikiKkQ}{500}
\newcommand{\NWikiKkCorpus}{30,000}
\newcommand{\NWikiUzQ}{500}
\newcommand{\NWikiUzCorpus}{30,000}
\newcommand{\NWikiKyQ}{500}
\newcommand{\NWikiKyCorpus}{20,587}

\newcommand{\NWikiUzLat}{499}
\newcommand{\NPoolKk}{166,263}
\newcommand{\NGroundedKk}{251,778}
\newcommand{\NDropHeldKk}{6,880}
\newcommand{\NPoolUz}{171,525}
\newcommand{\NGroundedUz}{179,142}
\newcommand{\NDropHeldUz}{7,612}
\newcommand{\NPoolKy}{99,293}
\newcommand{\NGroundedKy}{102,233}
\newcommand{\NDropHeldKy}{2,903}
\newcommand{\NDropTestKk}{69,404}
\newcommand{\NCaKk}{60,000}
\newcommand{\NCaUz}{60,000}
\newcommand{\NCaKy}{60,000}
\newcommand{\NCaTitleUz}{10,000}
\newcommand{\NCaTitleKy}{10,000}
\newcommand{\NCaTgLaw}{13,188}
\newcommand{\NCaTgTax}{257}
\newcommand{\NCaTgWeb}{14,996}
\newcommand{\NCaTgWiki}{5,280}
\newcommand{\NCaTgSvc}{3,935}
\newcommand{\NCaTotal}{279,904}
\newcommand{\NCaTajik}{37,656}
\newcommand{\NCaReplay}{42,248}

\newcommand{\NCaReplayEn}{24,993}
\newcommand{\NCaReplayRu}{17,255}
\newcommand{\PctCaTajik}{13}
\newcommand{\PctCaReplay}{15}
\newcommand{\PctCaNatural}{64}
\newcommand{\PctCaTitle}{7}
\newcommand{\NCaWithNeg}{219,399}
\newcommand{\PctCaWithNeg}{92}

\newcommand{\NSetKkLang}{166,263}

\newcommand{\NSetUzLang}{171,525}
\newcommand{\NSetUzTitle}{10,000}
\newcommand{\NSetReplay}{42,248}

\newcommand{\StepsKk}{410}
\newcommand{\StepsUz}{441}
\newcommand{\StepsCa}{562}
\newcommand{\CaSetsMatch}{identical}
\newcommand{\TrainLr}{$2\times10^{-5}$}
\newcommand{\CaMinutes}{181}
\newcommand{\CaLossFirst}{0.24}
\newcommand{\CaLossLastMean}{0.18}

\newcommand{\MiniBatch}{16}
\newcommand{\MiniBatchUz}{48}
\newcommand{\NLayers}{24}
\newcommand{\Vocab}{262,144}
\newcommand{\EmbDim}{768}
\newcommand{\QueryPrompt}{task: search result \textbar{} query: }
\newcommand{\DocPrompt}{title: none \textbar{} text: }

%% file: pack/macros_extra.tex
\newcommand{\ModelName}{Saidzoda-CA-EmbeddingGemma-300M}

%% file: sections/abstract.tex
The small open embedder EmbeddingGemma-300M, used as released, finds the right passage first for only \BaseBelTgRone\% to \BaseBelKkRone\% of Belebele questions in Tajik, Kazakh, Uzbek and Kyrgyz.
We present \ModelName, one model for all four languages.
It is our Tajik fine-tune of EmbeddingGemma-300M, trained for one more epoch on \NCaTotal{} pairs: questions in Kazakh, Uzbek and Kyrgyz that our own language model wrote about source passages, a Tajik refresher, and English and Russian replay.
On Belebele, recall@1 rises by \CaBelGainMin{} to \CaBelGainMax{} points in each language, to between \CaBelKkRone\% and \CaBelTgRone\%.
Against single-language models trained with the same settings, the one model stays within \CaVsKkBelKkRoneAbs{} points at recall@1 in Kazakh, Uzbek and Tajik, with overlapping recall@10 intervals, though it also had more training.
It keeps the Tajik law recall@10 of its starting point (\CaLawRten{} against \TgLawRten) but loses \CaVsTgLawRoneAbs{} points of law recall@1.
A model trained on one of these languages also improves the others on Belebele, but less than training on them.
The model still trails the 560M XLM-RoBERTa embedders on Kazakh, Uzbek and Kyrgyz Belebele: the better of multilingual-e5-large-instruct and BGE-M3, used as released, is \TurkicGapMin{} to \TurkicGapMax{} points ahead at recall@1.
Our scores of those released models agree with the published MTEB results within \MtebMaxDiff{} points.
Russian and English recall@1 fall by \CaVsBaseRuRoneAbs{} and \CaVsBaseEnRoneAbs{} points.

%% file: sections/intro.tex
\section{Introduction}

A retrieval system for Central Asia has to read four languages: Tajik, which is Persian written in Cyrillic, and three Turkic languages, Kazakh and Kyrgyz in Cyrillic and Uzbek mostly in Latin.
Training and serving one embedder per language multiplies the work by four.
We ask whether one small model can serve all four, and what it loses against models trained for one language.

Our earlier reports adapted language models to Tajik \citep{saidzoda2026lfm,saidzoda2026qwen3} and fine-tuned the 560M multilingual-e5-large-instruct \citep{wang2024me5} for Tajik retrieval \citep{saidzoda2026e5}.
A companion report, submitted to arXiv, fine-tunes the much smaller EmbeddingGemma-300M \citep{vera2025embeddinggemma} for Tajik \citep{saidzoda2026gemmatg}; that Tajik model is the starting point of this work.
Here we train it once more on Kazakh, Uzbek and Kyrgyz question and passage pairs, with a share of the Tajik pairs, and compare the result with the released model, with single-language Tajik, Kazakh and Uzbek models trained the same way, and with larger models.

EmbeddingGemma-300M was our choice for the small model because its tokenizer represents every Tajik-specific letter (we did not test the other alphabets letter by letter), it produces \EmbDim-dimensional vectors, and, used as released, it is weak in all four languages: on Belebele \citep{bandarkar2024belebele}, the one test in our set not written by our generator, its recall@1 is \BaseBelTgRone{} in Tajik, \BaseBelKkRone{} in Kazakh, \BaseBelUzRone{} in Uzbek and \BaseBelKyRone{} in Kyrgyz.

Our contributions are:
\begin{enumerate}
\item \textbf{One model for four languages.} \ModelName{} and its recipe: continued training of a Tajik model on generated Kazakh, Uzbek and Kyrgyz questions grounded in source passages with a Tajik refresher and replay (Sections~\ref{sec:data} and~\ref{sec:recipe}).
\item \textbf{One model against three single-language models.} Kazakh, Uzbek and Tajik models trained the same way show what the shared model gives up in each language, and how much a single-language model already transfers to its neighbours (Section~\ref{sec:results}).
\item \textbf{The gap to larger models.} The released 560M models stay \TurkicGapMin{} to \TurkicGapMax{} points ahead of our 300M model at recall@1 on the Turkic Belebele sets; our scores of those models agree with the published MTEB results (Section~\ref{sec:larger}).
\end{enumerate}

%% file: sections/languages.tex
\section{Four Languages, One Culture}
\label{sec:languages}

Tajik is Persian: Tajik, Iranian and Afghan Persian are mutually intelligible varieties of the same language, and Tajik is written in a modified Cyrillic alphabet \citep{merchant2025connecting}.
Kazakh, Uzbek and Kyrgyz are Turkic languages \citep{glottolog}.
The two groups share one culture, which historians call Turko-Persian \citep{green2019persianate}.
Under the Samanids, who ruled from Bukhara from the ninth century, Persian became a written language of state, and by the tenth century the court poetry of Bukhara had made it a language of government and learning in a region that later stretched from China to the Balkans.
The Turkic dynasties that ruled after the Samanids, from the Ghaznavids and Saljuqs to the Timurids and Shaybanids, kept it as their language of state.
Across much of the region Persian was ``a literate island in a sea of spoken Turkic'', the written language of courts and scholars while many nomadic peoples spoke Turkic; as \citet{green2019persianate} notes, a learned language of this kind is ``geographically broad but socially shallow''.
For centuries Bukhara, Samarkand, Tashkent and Kokand were also centres of Islamic learning, and their madrasas were still numerous on the eve of the Russian invasion: even at the end of the nineteenth century Samarkand province alone had fifty-eight, and the madrasas of Kokand attracted students even from Bukhara \citep{khalid1998politics}.
Writers of the time counted between 200 and 400 madrasas in the Emirate of Bukhara, whose students came from Iran, Afghanistan, India, Kashgar, Crimea, Tatarstan and the North Caucasus \citep{bobojonova2022madrasahs}; Bukhara was a common destination for Tatar scholars, and nomadic families hired teachers from Bukhara for their children \citep{khalid1998politics}.
Bedil, whose Persian poetry was widely read in Afghanistan and Transoxiana, was born in India to a family of the Arlas, a Chaghatay Turkic tribe \citep{ferhadi1992bidil}.

This long contact left a shared vocabulary.
The names of the days of the week in Uzbek, Kazakh and Kyrgyz come from Persian \citep{rosenfeld1996week}; Uzbek surnames are mostly of Arabic or Persian origin \citep{koziyeva2022anthroponyms}; and the Tajik of Bukhara has taken Uzbek words, suffixes and even the question particle \textit{mi}, while almost all of its speakers also speak Uzbek \citep{ido2014bukharan}.
The Russian Empire invaded the region in the 1860s and 1870s.
Under Soviet rule the madrasas were closed: by 1927 madrasa education had been abolished in Tashkent and cut sharply in Bukhara and Samarkand, before the assault on Islam of the 1930s.
The national-territorial delimitation of 1924 created national republics in place of Turkestan, Bukhara and Khiva, and the alphabet was changed twice, Uzbek for example moving from the Arabic script to Latin and, in 1940, to Cyrillic \citep{khalid2015making}.
The Soviet period also added the Russian endings \textit{-ov} and \textit{-ova} to these surnames, a custom that continues today \citep{koziyeva2022anthroponyms}.

\paragraph{Why one model for four languages.} This shared history is why we treat the four languages together, although earlier multilingual work groups Kazakh, Uzbek and Kyrgyz as Turkic and leaves Tajik out. To our knowledge, \ModelName{} is the first retrieval embedder trained and evaluated for Tajik, Kazakh, Uzbek and Kyrgyz together.

%% file: sections/related.tex
\section{Related Work}

\paragraph{Multilingual embedders.}
Most open multilingual embedders build on XLM-RoBERTa \citep{conneau2020unsupervised}, among them multilingual E5 \citep{wang2024me5}, BGE-M3 \citep{chen2024bgem3} and mGTE \citep{zhang2024mgte}; newer ones are derived from decoder language models, such as Qwen3-Embedding \citep{zhang2025qwen3emb} and EmbeddingGemma \citep{vera2025embeddinggemma}, a 300M encoder initialised from Gemma 3 \citep{gemmateam2025gemma3}.
MTEB and MMTEB \citep{muennighoff2023mteb,enevoldsen2025mmteb} measure them across many languages; the only retrieval task there that covers all four of our languages is Belebele used as retrieval.

\paragraph{Central Asian retrieval.}
Turkic Central Asian languages have gained resources in recent years \citep{veitsman2025recent}.
For Kazakh there is the KazQAD question answering and retrieval set \citep{yeshpanov2024kazqad} and a peer-reviewed benchmark for duplicate detection, similarity and retrieval \citep{tleubayeva2026kazakhtextduplicates}.
For Uzbek, a legal retrieval study fine-tunes a retriever on expert-written legal questions and also reports scores of released embedders, among them EmbeddingGemma-300M, on an Uzbek retrieval subset \citep{danielyan2026uzbek}.
For Kyrgyz we found no peer-reviewed passage retrieval benchmark; an open project fine-tunes E5 on Kyrgyz pairs written by a language model.\footnote{\url{https://github.com/askarbekkk/kyrgyz-embeddings}, not peer-reviewed.}
For Tajik, apart from our own reports, we know of no published embedder; the nearest retrieval work is on Persian written in the Perso-Arabic script \citep{zinvandi2025famteb,sarmadi2025hakim}.
The massively multilingual embedders above cover all four languages, and, as Section~\ref{sec:larger} shows, the 560M ones are strong in the three Turkic languages; apart from our own four-language E5 models (Section~\ref{sec:larger}), a literature search of November 2026 found no model fine-tuned for these four languages together, and no small one.
Model merging is an alternative to joint training for adding languages to a model \citep{alexandrov2024mitigating,kunz2025family}; we did not try it.

\paragraph{Training on generated questions.}
Training retrievers on questions written by a language model goes back to InPars, Promptagator and GPL \citep{bonifacio2022inpars,dai2022promptagator,wang2022gpl}, and has been used to build training data in many languages at once \citep{thakur2024swimir}.
Test sets built the same way may favour systems close to the generator \citep{rahmani2024synthetic,rahmani2025bias}, which is why we read our generated test sets as upper bounds.
We train with the in-batch multiple-negatives loss \citep{henderson2017smartreply} with gradient caching \citep{gao2021gradcache} in sentence-transformers \citep{reimers2019sbert}, with hard negatives mined as in NV-Retriever \citep{moreira2024nvretriever}.

%% file: sections/data.tex
\section{Data and Benchmark}
\label{sec:data}

\paragraph{Grounded question and passage pairs.}
For Kazakh, Uzbek and Kyrgyz, according to our data-generation records, our Central Asian fine-tune of a 27B Qwen model wrote questions about source passages (Wikipedia, encyclopedias, news and web text, depending on the language), and a second model, the released Qwen3.6-27B, checked that each question is answered by its passage and that the answer is correct; only questions that passed were kept, and the passage itself, not the generated answer, is the positive.
The two models are of the same family, so the check is not fully independent.
We call these questions grounded, to separate them from the title questions below; they are generated, not written by people.
This gave \NGroundedKk{} Kazakh, \NGroundedUz{} Uzbek and \NGroundedKy{} Kyrgyz checked rows.
Before any pair was built, Wikipedia articles for testing were frozen by a hash-based split, and every row about a held-out article was dropped (\NDropHeldKk, \NDropHeldUz{} and \NDropHeldKy{} rows); rows whose passage is in a test corpus were also dropped (\NDropTestKk{} in Kazakh; none in Uzbek and Kyrgyz, whose test corpora were built only from held-out articles and passages no question used).
Positives that share any 13-word sequence with a Belebele or replay test passage were dropped, as were questions equal to a test question.
This leaves \NPoolKk{} Kazakh, \NPoolUz{} Uzbek and \NPoolKy{} Kyrgyz pairs.
In addition, earlier title-based Uzbek and Kyrgyz pairs, whose question asks about the subject of a Wikipedia article, were available; we use them in small numbers.
The Uzbek text is in both scripts, and each question is in the script of its passage.

\paragraph{Training set of our model.}
Table~\ref{tab:data} lists the \NCaTotal{} pairs.
Each Turkic language contributes \NCaKk{} grounded pairs (\PctCaNatural\% of the set together), sampled at random from its pool, so that sampling in proportion to size does not turn the model into a Kazakh model.
We add \NCaTitleUz{} Uzbek and \NCaTitleKy{} Kyrgyz title pairs (\PctCaTitle\%), a quarter of each Tajik training set of our Tajik model (\NCaTajik{} pairs, \PctCaTajik\%), and half of its English and Russian replay sets (\NCaReplay{} pairs, \PctCaReplay\%) from Natural Questions \citep{kwiatkowski2019nq}, MIRACL \citep{zhang2023miracl} and a Russian hard-negative set (ru-HNP).
\NCaWithNeg{} of the non-replay pairs (\PctCaWithNeg\%) carry one hard negative, mined with the released BGE-M3 \citep{chen2024bgem3} from ranks 10 to 50 of the passage pool of its language, from a different article than the positive and with a score below 0.95 times that of the positive \citep{moreira2024nvretriever}.
The Tajik pairs include a small set of customer-service requests written by a language model in a simulation of our own customer-service line; this part is not released.

\input{pack/tables/T4_data}

\paragraph{Single-language models.}
For comparison we trained a Kazakh model on all \NSetKkLang{} Kazakh grounded pairs and an Uzbek model on all \NSetUzLang{} Uzbek grounded pairs plus \NSetUzTitle{} title pairs, each with \NSetReplay{} replay pairs and each starting from the released EmbeddingGemma-300M.
The Tajik model is the one of our companion report \citep{saidzoda2026gemmatg}, trained on the Tajik pairs of our E5 report \citep{saidzoda2026e5}.
We did not train a Kyrgyz model.

\paragraph{Test sets.}
\begin{itemize}
\item \textbf{Belebele} \citep{bandarkar2024belebele} in Tajik, Kazakh, Uzbek (Latin script) and Kyrgyz: \NBelTgQ{} questions per language; the dataset is fully parallel, so the four languages share one set of questions, each to be matched to its own passage among \NBelTgCorpus{} distinct passages in the same language, as in the Belebele retrieval task of MMTEB \citep{enevoldsen2025mmteb}. It is the only test not produced by our generator, so our main claims rest on it.
\item \textbf{Wikipedia} in Kazakh, Uzbek and Kyrgyz: \NWikiKkQ{} generated questions per language on held-out articles, searched over \NWikiKkCorpus, \NWikiUzCorpus{} and \NWikiKyCorpus{} passages. The Uzbek and Kyrgyz sets were frozen before any of these models was trained. \NWikiUzLat{} of the \NWikiUzQ{} Uzbek questions are in Latin script, so Cyrillic Uzbek is effectively untested.
\item \textbf{Tajik}: the law (\NLawQ{} questions over \NLawCorpus{} chunks), tax (\NTaxQ{} over \NTaxCorpus) and Wikipedia (\NWikiQ) sets of our E5 report, used here only to check that Tajik is kept.
\item \textbf{Russian and English Belebele} to check that the replay languages are kept.
\end{itemize}
We report recall@1 and recall@10 with 95\% bootstrap intervals for recall@10 (1,000 resamples over questions, with the passages fixed; Belebele has several questions per passage, so these intervals are somewhat too narrow); the result files hold no intervals for recall@1, and no per-question ranks, so we cannot run paired tests.
All models are scored with the same script in bfloat16, each with its own prompts: for EmbeddingGemma ``\QueryPrompt'' and ``\DocPrompt'', for the E5 models the web-search instruction of multilingual-e5-large-instruct on queries, and none for BGE-M3.

%% file: pack/tables/T4_data.tex
\begin{table}[t]\centering\footnotesize\setlength{\tabcolsep}{4pt}
\begin{tabular}{lr}
\toprule
Part & Pairs \\
\midrule
Kazakh, grounded questions & \NCaKk \\
Uzbek, grounded questions & \NCaUz \\
Kyrgyz, grounded questions & \NCaKy \\
Uzbek, title questions & \NCaTitleUz \\
Kyrgyz, title questions & \NCaTitleKy \\
\midrule
Tajik law & \NCaTgLaw \\
Tajik tax & \NCaTgTax \\
Tajik Wikipedia & \NCaTgWiki \\
Tajik web and literature & \NCaTgWeb \\
Customer service (synthetic, Tajik) & \NCaTgSvc \\
\midrule
English replay & \NCaReplayEn \\
Russian replay & \NCaReplayRu \\
\midrule
Total & \NCaTotal \\
\bottomrule
\end{tabular}
\caption{Training pairs of our four-language model, as counted by the training run.}
\label{tab:data}
\end{table}

%% file: sections/recipe.tex
\section{Training Recipe}
\label{sec:recipe}

\paragraph{Starting point.}
Our model starts from our Tajik EmbeddingGemma model \citep{saidzoda2026gemmatg}, not from the released one, so that the Tajik ability does not have to be learned again; the Tajik share of the training set is there to keep it.
EmbeddingGemma-300M has a Gemma 3 encoder with \NLayers{} layers and a vocabulary of \Vocab{} tokens, and gives \EmbDim-dimensional vectors.
Its tokenizer maps none of the \GlyphCount{} Tajik-specific letter forms to the unknown token, and it splits a word into \FertTg{} (Tajik), \FertKk{} (Kazakh), \FertUz{} (Uzbek) and \FertKy{} (Kyrgyz) tokens on average, against \FertRu{} for Russian.

\paragraph{Settings.}
The recipe is the one of our Tajik model, with the Tajik model as the starting point instead of the released one: full fine-tuning with the cached multiple-negatives ranking loss (in-batch negatives plus the mined negative, scale 20), global batch 512, learning rate \TrainLr{} with linear decay and 5\% warm-up, one epoch, sequences up to 512 tokens, seed 42, fp32 with eager attention, and the model's own query and document prompts.
Each batch is drawn from one part of the training set, parts are sampled in proportion to their size, and no text appears twice in a batch.
The epoch took \StepsCa{} steps and about \CaMinutes{} minutes on one H100 GPU shared with another job, with a gradient-caching mini-batch of \MiniBatch.
The training pairs as counted by the run are \CaSetsMatch{} to Table~\ref{tab:data}.

The Kazakh and Uzbek models use the same settings from the released model, for \StepsKk{} and \StepsUz{} steps; the Uzbek model ran on a free card with a mini-batch of \MiniBatchUz, which does not change the global batch.
The Kazakh run was stopped by an operator error after about 165 steps and resumed from its step-100 checkpoint; the logged loss after the resume matched the stopped run.

\begin{figure}[t]\centering
\includegraphics[width=0.9\columnwidth]{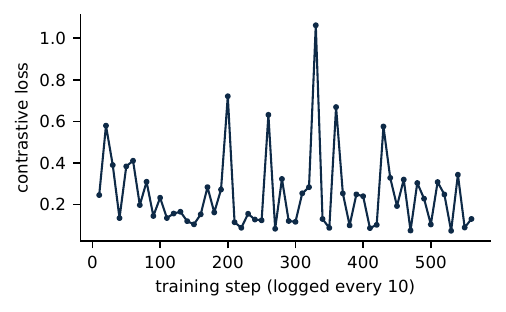}
\caption{Training loss of \ModelName, logged every 10 steps.}
\label{fig:loss}
\end{figure}

\paragraph{Training loss.}
The logged loss of our model is that of one batch every 10 steps; because each batch comes from one part of the training set, it varies widely from step to step (Figure~\ref{fig:loss}). It is \CaLossFirst{} at the first logged step and \CaLossLastMean{} on average over the last five, with no sign of divergence.

%% file: sections/results.tex
\section{Results}
\label{sec:results}

\input{pack/tables/T1_belebele}

\begin{figure}[t]\centering
\includegraphics[width=\columnwidth]{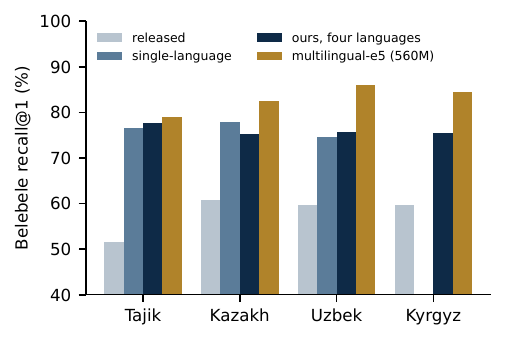}
\caption{Belebele recall@1 per language: EmbeddingGemma-300M as released, the single-language model for that language (none for Kyrgyz), our four-language model, and multilingual-e5-large-instruct as released.}
\label{fig:bel}
\end{figure}

\paragraph{Gains over the released model.}
Table~\ref{tab:belebele} gives Belebele recall@1 and Figure~\ref{fig:bel} draws it.
Our model gains \CaVsBaseBelTgRoneAbs{} points in Tajik, \CaVsBaseBelKkRoneAbs{} in Kazakh, \CaVsBaseBelUzRoneAbs{} in Uzbek and \CaVsBaseBelKyRoneAbs{} in Kyrgyz over EmbeddingGemma-300M as released, and the recall@10 intervals of the two models do not overlap in any of the four languages.
The mean over the four languages rises from \BaseBelMeanRone{} to \CaBelMeanRone.
On the generated Wikipedia sets (Table~\ref{tab:wiki}) recall@1 rises by \CaVsBaseWikiKkRoneAbs{} points in Kazakh, \CaVsBaseWikiUzRoneAbs{} in Uzbek and \CaVsBaseWikiKyRoneAbs{} in Kyrgyz; these sets start higher and their absolute values are optimistic.

\paragraph{One model against single-language models.}
In each language that has a single-language model, our model comes close to it.
In Kazakh it is \CaVsKkBelKkRoneAbs{} points below the Kazakh model at recall@1 (\CaBelKkRone{} against \KkBelKkRone), with overlapping recall@10 intervals; in Uzbek it is \CaVsUzBelUzRoneAbs{} points above the Uzbek model (\CaBelUzRone{} against \UzBelUzRone), and in Tajik \CaVsTgBelTgRoneAbs{} points above the Tajik model (\CaBelTgRone{} against \TgBelTgRone), both with overlapping intervals.
The Wikipedia sets show the same: \CaWikiKkRone{} against \KkWikiKkRone{} in Kazakh and \CaWikiUzRone{} against \UzWikiUzRone{} in Uzbek at recall@1, with overlapping recall@10 intervals.
So, within the resolution of these tests, one model does about as well as the three single-language models we trained, and also covers Kyrgyz, where it reaches \CaBelKyRone{} on Belebele.
The resolution is limited: we have one seed per model, no paired test and no interval for recall@1, so the Kazakh difference of \CaVsKkBelKkRoneAbs{} points may be real and smaller differences should not be read either way.
The comparison also favours our model: it starts from the Tajik model and so had two epochs of training in total, against one for the Kazakh and Uzbek models from the released model. We did not train the four-language mix from the released model, so we cannot say how much of the result is due to joint training itself.

\input{pack/tables/T2_wiki}

\paragraph{Effects outside the trained language.}
The single-language models also change the languages they were not trained on.
The Kazakh model raises Kyrgyz Belebele recall@1 from \BaseBelKyRone{} to \KkBelKyRone{} and Uzbek to \KkBelUzRone; the Uzbek model raises Kazakh to \UzBelKkRone{} and Kyrgyz to \UzBelKyRone; and the Tajik model, trained on Tajik with English and Russian replay only, raises the three Turkic languages by \TgVsBaseBelKkRoneAbs, \TgVsBaseBelUzRoneAbs{} and \TgVsBaseBelKyRoneAbs{} points.
The Kazakh and Uzbek models do not improve Tajik Belebele (\KkBelTgRone{} and \UzBelTgRone, against \BaseBelTgRone{} for the released model), although they gain \KkVsBaseLawRtenAbs{} and \UzVsBaseLawRtenAbs{} points of Tajik law recall@10, far short of the Tajik model's \TgLawRten.
In Kyrgyz the best single-language model reaches \KkBelKyRone, while our model, which saw Kyrgyz pairs, reaches \CaBelKyRone.
Since Tajik is not a Turkic language and the Tajik model still improves the Turkic sets, much of this effect may be general retrieval training, or adaptation to the Belebele format, rather than language similarity; we did not separate the two.

\paragraph{Tajik: recall@10 is kept, recall@1 drops.}
\input{pack/tables/T3_tajik}
Table~\ref{tab:tajik} compares our model with the Tajik model it started from.
On Tajik law recall@10 is \CaLawRten{} against \TgLawRten, on Wikipedia \CaWikiRten{} against \TgWikiRten, and on Tajik Belebele \CaBelTgRten{} against \TgBelTgRten, all with overlapping intervals.
Tax falls by \CaVsTgTaxRtenAbs{} points, also with overlapping intervals on its \NTaxQ{} questions.
At recall@1 law is \CaVsTgLawRoneAbs{} points lower (\CaLawRone{} against \TgLawRone{} on \NLawQ{} questions, a point estimate without interval) and Tajik Wikipedia \CaVsTgWikiRoneAbs{} points lower, although law, tax and Wikipedia pairs are part of the Tajik refresher, so these are in-domain tests.
Russian and English recall@10 stay high (\CaRuRten{} and \CaEnRten), but recall@1 is \CaVsBaseRuRoneAbs{} and \CaVsBaseEnRoneAbs{} points below the released model; most of the Russian loss was already present in the Tajik model.

\paragraph{Against larger models.}
\label{sec:larger}
The lower block of Table~\ref{tab:belebele} gives the 560M models.
Used as released, multilingual-e5-large-instruct and BGE-M3 are ahead of our model on Kazakh, Uzbek and Kyrgyz Belebele: the better of the two leads by \TurkicGapMin{} to \TurkicGapMax{} points at recall@1, and the recall@10 intervals of our model and multilingual-e5-large-instruct do not overlap in these three languages.
In Tajik the picture is different: our model is close to multilingual-e5-large-instruct on Belebele (\CaBelTgRten{} against \EfiveBelTgRten{} recall@10, overlapping) and well ahead on law (\CaLawRten{} against \EfiveLawRten), because the released 560M models are weak on Tajik domain text \citep{saidzoda2026e5}.
On the generated Wikipedia sets our model is close to multilingual-e5-large-instruct in the three Turkic languages (recall@1 \CaWikiKkRone, \CaWikiUzRone{} and \CaWikiKyRone{} against \EfiveWikiKkRone, \EfiveWikiUzRone{} and \EfiveWikiKyRone), but these sets come from the same generator as our training questions and are near the ceiling at recall@10, so they say little about the gap.
Our own four-language E5 models, which will be described in a separate report, reach Belebele means of \ArmABelMeanRone{} and \ArmBBelMeanRone{} against \CaBelMeanRone{} for our model.
We did not compare with released embedders of our model's size.

To check our scoring, we compared it with the published MTEB results of the same three released models on the same-language Belebele retrieval subsets (Table~\ref{tab:mteb}).\footnote{\url{https://github.com/embeddings-benchmark/results}, files \texttt{BelebeleRetrieval.json}, read 6 November 2026.}
Our recall@1 values for the three released models are within \MtebMaxDiff{} points of the published ones in all twelve model and language pairs; we did not trace the remaining differences, which may come from scoring details such as precision, prompts or library versions.
This checks our scoring of the released models, not our model, which is not in MTEB.

\input{pack/tables/T5_mteb}

%% file: pack/tables/T1_belebele.tex
\begin{table*}[t]\centering\small
\begin{tabular}{lrrrrr}
\toprule
Model & Tajik & Kazakh & Uzbek & Kyrgyz & Mean \\
\midrule
\multicolumn{6}{l}{\emph{EmbeddingGemma-300M, as released and fine-tuned by us}} \\
EmbeddingGemma-300M & \BaseBelTgRone & \BaseBelKkRone & \BaseBelUzRone & \BaseBelKyRone & \BaseBelMeanRone \\
Tajik model & \TgBelTgRone & \TgBelKkRone & \TgBelUzRone & \TgBelKyRone & \TgBelMeanRone \\
Kazakh model & \KkBelTgRone & \textbf{\KkBelKkRone} & \KkBelUzRone & \KkBelKyRone & \KkBelMeanRone \\
Uzbek model & \UzBelTgRone & \UzBelKkRone & \UzBelUzRone & \UzBelKyRone & \UzBelMeanRone \\
Ours (four languages) & \textbf{\CaBelTgRone} & \CaBelKkRone & \textbf{\CaBelUzRone} & \textbf{\CaBelKyRone} & \CaBelMeanRone \\
\midrule
\multicolumn{6}{l}{\emph{560M models, for reference}} \\
multilingual-e5-large-instruct & \EfiveBelTgRone & \EfiveBelKkRone & \EfiveBelUzRone & \EfiveBelKyRone & \EfiveBelMeanRone \\
BGE-M3 & \BgeBelTgRone & \BgeBelKkRone & \BgeBelUzRone & \BgeBelKyRone & \BgeBelMeanRone \\
Saidzoda-Tajik-E5-Large & \EfTgBelTgRone & \EfTgBelKkRone & \EfTgBelUzRone & \EfTgBelKyRone & \EfTgBelMeanRone \\
E5, four languages, arm A & \ArmABelTgRone & \ArmABelKkRone & \ArmABelUzRone & \ArmABelKyRone & \ArmABelMeanRone \\
E5, four languages, arm B & \ArmBBelTgRone & \ArmBBelKkRone & \ArmBBelUzRone & \ArmBBelKyRone & \ArmBBelMeanRone \\
\bottomrule
\end{tabular}
\caption{Belebele retrieval, recall@1 in percent: each of the \NBelTgQ{} questions per language (the same parallel questions in every language) must find its own passage among \NBelTgCorpus{} passages in the same language. Bold marks the highest value among the five EmbeddingGemma rows, not a significant difference. The Tajik, Kazakh and Uzbek models are our single-language fine-tunes; no Kyrgyz single-language model was trained. The two E5 arms are our four-language E5 models, which will be described in a separate report.}
\label{tab:belebele}
\end{table*}

%% file: pack/tables/T2_wiki.tex
\begin{table*}[t]\centering\small
\begin{tabular}{lrrrrrr}
\toprule
& \multicolumn{2}{c}{Kazakh (\NWikiKkQ)} & \multicolumn{2}{c}{Uzbek (\NWikiUzQ)} & \multicolumn{2}{c}{Kyrgyz (\NWikiKyQ)} \\
Model & R@1 & R@10 & R@1 & R@10 & R@1 & R@10 \\
\midrule
EmbeddingGemma-300M & \BaseWikiKkRone & \BaseWikiKkRten & \BaseWikiUzRone & \BaseWikiUzRten & \BaseWikiKyRone & \BaseWikiKyRten \\
Tajik model & \TgWikiKkRone & \TgWikiKkRten & \TgWikiUzRone & \TgWikiUzRten & \TgWikiKyRone & \TgWikiKyRten \\
Kazakh model & \KkWikiKkRone & \KkWikiKkRten & \KkWikiUzRone & \KkWikiUzRten & \KkWikiKyRone & \KkWikiKyRten \\
Uzbek model & \UzWikiKkRone & \UzWikiKkRten & \UzWikiUzRone & \UzWikiUzRten & \UzWikiKyRone & \UzWikiKyRten \\
Ours (four languages) & \CaWikiKkRone & \CaWikiKkRten & \CaWikiUzRone & \CaWikiUzRten & \CaWikiKyRone & \CaWikiKyRten \\
\midrule
multilingual-e5-large-instruct & \EfiveWikiKkRone & \EfiveWikiKkRten & \EfiveWikiUzRone & \EfiveWikiUzRten & \EfiveWikiKyRone & \EfiveWikiKyRten \\
E5, four languages, arm B & \ArmBWikiKkRone & \ArmBWikiKkRten & \ArmBWikiUzRone & \ArmBWikiUzRten & \ArmBWikiKyRone & \ArmBWikiKyRten \\
\bottomrule
\end{tabular}
\caption{Wikipedia questions on held-out articles (recall@1 and recall@10, percent), written by the same generator family as the training questions, so the absolute values are optimistic. Corpora: \NWikiKkCorpus{} (Kazakh), \NWikiUzCorpus{} (Uzbek) and \NWikiKyCorpus{} (Kyrgyz) passages. BGE-M3 was not scored on these sets.}
\label{tab:wiki}
\end{table*}

%% file: pack/tables/T3_tajik.tex
\begin{table}[t]\centering\footnotesize\setlength{\tabcolsep}{3.5pt}
\begin{tabular}{lrrr}
\toprule
Test set & Released & Tajik & Ours \\
\midrule
\multicolumn{4}{l}{\emph{Recall@10}} \\
Tajik law & \BaseLawRten & \TgLawRten & \CaLawRten \\
Tajik tax & \BaseTaxRten & \TgTaxRten & \CaTaxRten \\
Tajik Wikipedia & \BaseWikiRten & \TgWikiRten & \CaWikiRten \\
Belebele Tajik & \BaseBelTgRten & \TgBelTgRten & \CaBelTgRten \\
Belebele Russian & \BaseRuRten & \TgRuRten & \CaRuRten \\
Belebele English & \BaseEnRten & \TgEnRten & \CaEnRten \\
\midrule
\multicolumn{4}{l}{\emph{Recall@1}} \\
Tajik law & \BaseLawRone & \TgLawRone & \CaLawRone \\
Belebele Tajik & \BaseBelTgRone & \TgBelTgRone & \CaBelTgRone \\
Belebele Russian & \BaseRuRone & \TgRuRone & \CaRuRone \\
Belebele English & \BaseEnRone & \TgEnRone & \CaEnRone \\
\bottomrule
\end{tabular}
\caption{Tajik and the replay languages (percent): EmbeddingGemma-300M as released, our Tajik model (the starting point of our model), and our four-language model. Law: \NLawQ{} questions; tax: \NTaxQ.}
\label{tab:tajik}
\end{table}

%% file: pack/tables/T5_mteb.tex
\begin{table}[t]\centering\footnotesize\setlength{\tabcolsep}{3pt}
\begin{tabular}{lrrrr}
\toprule
Model & tg & kk & uz & ky \\
\midrule
EmbeddingGemma-300M, MTEB & \MtebBaseBelTgRone & \MtebBaseBelKkRone & \MtebBaseBelUzRone & \MtebBaseBelKyRone \\
\quad ours & \BaseBelTgRone & \BaseBelKkRone & \BaseBelUzRone & \BaseBelKyRone \\
mE5-large-instruct, MTEB & \MtebEfiveBelTgRone & \MtebEfiveBelKkRone & \MtebEfiveBelUzRone & \MtebEfiveBelKyRone \\
\quad ours & \EfiveBelTgRone & \EfiveBelKkRone & \EfiveBelUzRone & \EfiveBelKyRone \\
BGE-M3, MTEB & \MtebBgeBelTgRone & \MtebBgeBelKkRone & \MtebBgeBelUzRone & \MtebBgeBelKyRone \\
\quad ours & \BgeBelTgRone & \BgeBelKkRone & \BgeBelUzRone & \BgeBelKyRone \\
\bottomrule
\end{tabular}
\caption{Recall@1 (percent) of the three released models on the same-language Belebele retrieval subsets as published in the MTEB results repository (MTEB) and as scored by us (ours). mE5 is multilingual-e5-large-instruct.}
\label{tab:mteb}
\end{table}

%% file: sections/discussion.tex
\section{Discussion}
\label{sec:discussion}

\paragraph{When to use which model.}
Where one small model has to serve all four languages, for example on a modest server or with a small index, our model gives each language close to what a model trained for that language alone gives, and keeps the Tajik domain recall@10 of our Tajik model, with some loss at recall@1.
Where quality in Kazakh, Uzbek or Kyrgyz matters most and a 560M model fits, a released XLM-RoBERTa embedder is the better choice for reading-style questions, as the Belebele scores show.
For Tajik domain text both released 560M models are weak \citep{saidzoda2026e5}, so a Tajik fine-tune, ours or the E5 one, is the better choice there.

\paragraph{What we cannot explain yet.}
Fine-tuning closes most of the Tajik gap between EmbeddingGemma-300M and the 560M models, but not the Turkic gap on Belebele.
The Turkic pairs are generated questions on Wikipedia, encyclopedia, news and web passages, while Belebele asks reading-comprehension questions about short news and travel passages; the smaller model, the different pretraining and this difference in question style are all candidates, and we did not test them.
We also did not test a larger share of Turkic pairs, a second epoch, or training from the released model instead of the Tajik one.

\paragraph{Costs.}
The model loses \CaVsBaseRuRoneAbs{} and \CaVsBaseEnRoneAbs{} points of Russian and English recall@1 against the released model, although \PctCaReplay\% of its pairs are English and Russian replay; we did not train without replay or with more of it.
In earlier experiments with the larger E5 model, continuing our Tajik E5 model on Central Asian pairs whose Uzbek and Kyrgyz part was title-based (questions about the subject of an article) lowered Uzbek and Kyrgyz Belebele recall@1 (\EfTgBelUzRone{} to \ArmABelUzRone{} and \EfTgBelKyRone{} to \ArmABelKyRone), while Kazakh and Tajik moved by about one point (\EfTgBelKkRone{} to \ArmABelKkRone{} and \EfTgBelTgRone{} to \ArmABelTgRone).
We guessed that the question style was the cause, though that run differed in other ways too, so here the title pairs are only \PctCaTitle\% of the set and the rest are grounded questions; we did not run this model without them.

%% file: sections/release.tex
\section{Release}
\label{sec:release}

\paragraph{What we release.}
The model is on the Hugging Face Hub as \texttt{Tohirju/\allowbreak\ModelName} in sentence-transformers format with its prompts, together with its training log and the result files behind every score in this report.
Access is given on request: the repository is public, and each request is approved by us.
The training pairs and the code are available on request; the customer-service pairs are not released.

\paragraph{Licence.}
EmbeddingGemma-300M is distributed under the Gemma Terms of Use (\texttt{ai.google.dev/gemma/terms}), and our weights are a model derivative under those terms: the repository states that the weights were modified, includes the Terms and the Gemma Prohibited Use Policy, and passes their use restrictions on to every user, and the access gate requires accepting them.
In addition, our model is offered under the Saidzoda model licence: organisations with revenue need a paid licence, while startups and individuals without revenue may use it free of charge until they earn from it; where the two conflict, the Gemma Terms govern.
Access requests go through \texttt{ai.saidzoda.com/access}.

\paragraph{How to use it.}
Load the model with sentence-transformers, encode questions with the ``query'' prompt and passages with the ``document'' prompt, and rank passages by cosine similarity, with inputs of up to 512 tokens.
The language of a question does not need to be given.
Retrieved text should be checked before an answer reaches a user.

%% file: sections/conclusion.tex
\section{Conclusion}

One epoch of continued training on generated Kazakh, Uzbek and Kyrgyz questions grounded in source passages, with a Tajik refresher and replay, turned our Tajik EmbeddingGemma-300M into one model for four Central Asian languages.
It gains \CaBelGainMin{} to \CaBelGainMax{} points of Belebele recall@1 over the released model, comes within \CaVsKkBelKkRoneAbs{} points of single-language models in each language where we have one, and keeps Tajik law recall@10.
It does not reach the 560M XLM-RoBERTa embedders on Kazakh, Uzbek and Kyrgyz Belebele, where they lead by \TurkicGapMin{} to \TurkicGapMax{} points; our scores of those models agree with the published MTEB results.
The missing evaluations are questions written by speakers of each language about real documents, and Cyrillic Uzbek.

%% file: sections/limitations.tex
\section*{Limitations}

\paragraph{Generated test questions.}
The Wikipedia questions in all three Turkic languages, and the Tajik law, tax and Wikipedia questions, were written by language models of the same family as the training questions, on disjoint articles and documents, so their scores are upper bounds.
Belebele is independent, but it consists of reading-comprehension questions about short news and travel texts, not the questions people ask a search system, and it is fully parallel across languages, so the four languages are not independent samples.

\paragraph{Statistics.}
The intervals are per-model bootstrap intervals for recall@10 that resample questions only; we have no paired tests and no intervals for recall@1, and most of our comparisons are at recall@1. Gains of more than ten points are safe; differences of a few points are not.
Every model was trained once, with one seed.
The tax set has only \NTaxQ{} questions.

\paragraph{Confounds.}
Our model starts from the Tajik model and the Kazakh and Uzbek models start from the released one, so our model had two epochs of training in total against one, and the comparison mixes the effect of joint training with that of the starting point and the extra training.
The Kazakh and Uzbek models used all their grounded pairs, while our model used a capped sample of each; the Uzbek model ran with a larger gradient-caching mini-batch on a free card.
The negatives were mined with the released BGE-M3.

\paragraph{Scope.}
We have no Kyrgyz single-language model, and nearly all Uzbek test questions are in Latin script, so Cyrillic Uzbek is untested.
We compared only with 560M released embedders, not with released models of our model's size.
We tested retrieval only, with inputs of up to 512 tokens, in bfloat16 scoring.
We report no latency: the timing in our result files was measured on shared, busy cards.
The customer-service training pairs are not released, so that part of the training data cannot be reproduced by others.

%% file: sections/disclosure.tex
\section*{Use of AI Assistants}

We used AI assistants (Anthropic Claude models) to write data-processing, training and analysis scripts, to search for and check references, and to draft parts of the text.
The training and Wikipedia test questions were generated by our own language models, as described in Section~\ref{sec:data}.
The authors designed the study, ran the training, checked every number against the recorded files, and take full responsibility for the content.
Every number in this report is generated by a script from logged files, and each reference was checked against its arXiv, ACL Anthology or DOI record.
AI assistants are not authors.

\section*{Acknowledgements}

We thank NVIDIA for access to the NVIDIA Inception programme, through which the training compute for this work was provided.

%% file: pack/tables/T6_examples.tex
\begin{table*}[t]\centering\small
\begin{tabular}{lp{0.5\textwidth}p{0.33\textwidth}}
\toprule
Set & Question & English gloss \\
\midrule
Kazakh & \tj{Томск облысы мен Краснояр өлкесі жер аумақтарынан өтетін Индыке өзенінің жалпы ұзындығы қанша}? & What is the total length of the Indyke river, which flows through Tomsk Oblast and Krasnoyarsk Krai? \\
Uzbek & Pol Genri Dyuks qayerda va qachon tavallud topgan? & Where and when was Paul Henry Dukes born? \\
Kyrgyz & 1648-\tj{жылдагы Вестфаль тынчтыгы кайсы согушту аяктаган жана кайсы эки шаарда түзүлгөн}? & Which war did the Peace of Westphalia of 1648 end, and in which two cities was it concluded? \\
\bottomrule
\end{tabular}
\caption{Example test questions from the Wikipedia sets, exactly as written by the generator. The English glosses are ours.}
\label{tab:examples}
\end{table*}